\documentclass[10pt,twocolumn,letterpaper]{article}

\usepackage[pagenumbers]{cvpr} 
\usepackage{multirow}
\usepackage[dvipsnames]{xcolor}

\definecolor{cvprblue}{rgb}{0.21,0.49,0.74}
\usepackage[pagebackref,breaklinks,colorlinks,citecolor=cvprblue]{hyperref}

\def\paperID{4767} 
\def\confName{CVPR}
\def\confYear{2024}
\DeclareMathOperator*{\argmax}{arg\,max}
\title{Context-based and Diversity-driven Specificity in Compositional Zero-Shot Learning}

\author{Yun Li\\
CSIRO's Data61\\
{\tt\small y.li@csiro.au}
\and
Zhe Liu\\
Bytedance Ltd.\\
{\tt\small zhe.liu01@bytedance.com}
\and
Hang Chen\\
Snap Inc.\\
{\tt\small chenhang386@gmail.com}
\and
Lina Yao\\
CSIRO's Data61\\
{\tt\small lina.yao@data61.csiro.au}
}

\begin{document}
\maketitle
\begin{abstract}

  Compositional Zero-Shot Learning (CZSL) aims to recognize unseen attribute-object pairs based on a limited set of observed examples. Current CZSL methodologies, despite their advancements, tend to neglect the distinct specificity levels present in attributes. For instance, given images of sliced strawberries, they may fail to prioritize `Sliced-Strawberry' over a generic `Red-Strawberry', despite the former being more informative. They also suffer from ballooning search space when shifting from Close-World (CW) to Open-World (OW) CZSL. To address the issues, we introduce the Context-based and Diversity-driven Specificity learning framework for CZSL (CDS-CZSL). Our framework evaluates the specificity of attributes by considering the diversity of objects they apply to and their related context. This novel approach allows for more accurate predictions by emphasizing specific attribute-object pairs and improves composition filtering in OW-CZSL. We conduct experiments in both CW and OW scenarios, and our model achieves state-of-the-art results across three datasets.

\end{abstract}    
\section{Introduction}
\label{sec:intro}


Humans effortlessly combine known ideas, such as the \textit{pink} of a rose and a blue \textit{dolphin}, to recognize unseen concepts like a \textit{pink} \textit{dolphin}.
This ability to learn compositionally is a hallmark of human intelligence~\cite{lake2014towards}, enabling us to infer vast knowledge from limited primitives without seeing every possible combination.
Inspired by this capability, Compositional Zero-Shot Learning (CZSL)~\cite{li2020symmetry,naeem2021learning,karthik2022kg} emerged. In CZSL, the goal is to train models on images of seen attribute-object pairs (e.g., attributes like colors and objects like animals) so they can recognize unseen pairs, thereby minimizing the need for extensive training datasets.



\begin{figure}
\small
  \centering
  \begin{subfigure}{0.27\linewidth}
  \centering
  \includegraphics[width=0.62\textwidth]{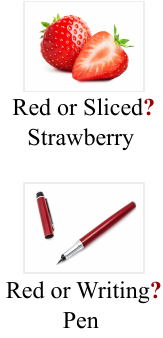}
    \caption{Which is better?}
    \label{fig:intro-a}
  \end{subfigure}
  \quad
  \centering
  \begin{subfigure}{0.65\linewidth}
  \includegraphics[width=\textwidth]{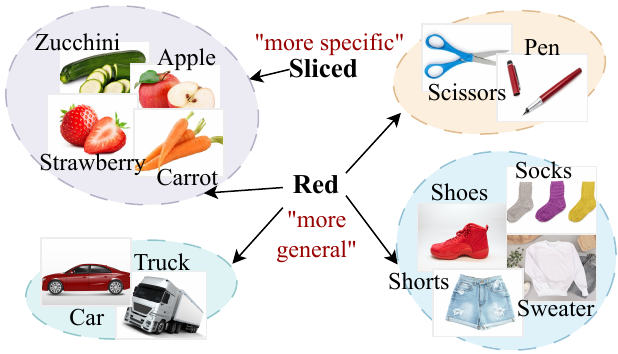}
    \caption{Diversity-driven specificity learning.}
    \label{fig:intro-b}
  \end{subfigure}
  \caption{Specificity in CZSL. (a)For strawberries, \textit{Sliced} is more specific than \textit{Red}. Instead, \textit{Red} is more specific than \textit{Writing} for a pen, as \textit{Writing} is its inherent function. 
  (b) Clustering images based on their object features, \textit{Red} spans multiple object clusters. In contrast, \textit{Sliced}, though applicable to several objects, only links to the food cluster, indicating its greater specificity.}
  \label{fig:intro}
\end{figure}

Traditional CZSL methods often adopt one of two strategies: 1) they project attribute-object textual labels and images into a shared space for direct, similarity-based composition classification~\cite{xu2021zero,naeem2021learning,wei2019adversarial}; 2) they use dual modules to classify attributes and objects separately, later fusing the results for the final composition~\cite{karthik2022kg,liu2023pami,li2023distilled}. 
Recently, their performance has been further improved by integrating large, pre-trained vision-language models like CLIP~\cite{radford2021learning}. Harnessing the powerful visual-semantic aligning capabilities of CLIP, these methods~\cite{csp2023, wang2023hierarchical,lu2023decomposed} set new performance benchmarks that surpass traditional approaches.

While current CZSL methods show promise, they often focus on optimizing model performance while overlooking an inherent challenge in CZSL: specificity in attributes. Unlike objects, which generally have straightforward definitions, attributes can be multifaceted. Consider the example in \cref{fig:intro-a}: a strawberry can be described as \textit{Red}, a common descriptor, or \textit{Sliced}, which is more specific. The value of such specific descriptors is supported by the Shannon Theory, which holds that rarer events offer more information than common ones ~\cite{shannon1948mathematical}. However, in pursuit of overall accuracy and broader generalization to unseen pairs, CZSL models may favor general attributes like \textit{Red}, which have wider applicability to almost all categories, over more specific yet valid descriptors like \textit{Sliced}. The challenge of effectively prioritizing these specific attributes without compromising accuracy remains under-explored.

Beyond the challenge of specificity, transitioning from a Closed-World (CW) setting~\cite{purushwalkam2019task,nagarajan2018attributes,li2020symmetry} to an Open-World (OW) setting~\cite{karthik2022kg,mancini2021open,mancini2022learning} poses additional difficulties. In CW, test set compositions are predefined and given as prior knowledge, making it less realistic for real-world applications. The OW setting, free of such restrictions, faces a largely expanded output space. Direct composition classification methods, as a result, suffer significant performance drops in OW-CZSL~\cite{mancini2021open}. To address this, some techniques narrow the search space by determining composition feasibility based on their similarity to seen pairs~\cite{lu2023decomposed,mancini2022learning,csp2023}. However, similarity-based feasibility estimation risks discarding specific attribute-object pairs in favor of more generic compositions.
In contrast, methods that separately predict attributes and objects can avoid this expanding search space and achieve advanced learning of primitives, i.e., attributes and objects~\cite{karthik2021revisiting,liu2023pami,li2023distilled}. Yet, these methods might not grasp the contextual nuances present in the composition space, like how \textit{Small} appears differently in contexts such as Small-Cat versus Small-House.

To address the aforementioned challenges in CZSL, we introduce the Context-based and Diversity-driven Specificity learning framework for CZSL (CDS-CZSL). Our framework employs a 3-branch structure: a composition-wise branch for contextual understanding through composition classification and two adapter-enhanced primitive-wise branches to extract attribute and object features effectively.

A distinguishing feature of our framework is the introduction of a context-based and diversity-driven specificity learner. Intuitively, the specificity of an attribute is linked to the range of objects it can describe. We thus cluster object features and drive the learning of an attribute's specificity using the diversity of clusters it covers, as depicted in \cref{fig:intro-b}. We further incorporate the context of attributes into specificity learning. This is rooted in the observation that the specificity of an attribute can vary depending on the object it's paired with. For instance, while \textit{Red} is general for strawberries, it is more specific than the inherent function attribute \textit{Writing} for pen, as illustrated in \cref{fig:intro-a}. Armed with the specificity insights, we refine attribute predictions to emphasize specific pairs. 

This specificity also aids in pruning the composition space, filtering out both overly specific and overly generic pairs, alleviating the challenges of composition space explosion in the OW setting. Crucially, using specificity, our method can retain valid, specific pairs that other feasibility calibration techniques~\cite{lu2023decomposed} might overlook.

In summary, our primary contributions include:

\noindent 1) We propose CDS-CZSL, a novel CLIP-based CZSL method that employs a 3-branch structure to equip both contextual understanding and efficient attribute/object learning.

\noindent 2) We introduce the specificity concept into attribute predictions for CZSL. The proposed specificity learner prioritizes specific attributes with accuracy holds. This specificity further enhances composition filtering in OW-CZSL setting.

\noindent 3) Our model achieves state-of-the-art (SOTA) results on three benchmark datasets in both CW and OW scenarios.

\section{Related Work}
\label{sec:relaetd}







\paragraph{Compositional Zero-shot Learning (CZSL).}

CZSL~\cite{naeem20223d,yang2022decomposable,hou2021detecting,kato2018compositional,liu2023pami,nagarajan2018attributes} has two main strategies for inferring unseen compositions. The first assumes that unseen and seen compositions share the same attribute and object scopes. Thus, this strategy first predicts primitive labels and then combines them to obtain the composition label~\cite{naeem2021learning,purushwalkam2019task,nagarajan2018attributes,li2020symmetry,liu2023pami}. For example,  Liu \etal~\cite{liu2023pami} separately predict attributes and objects based on contextual semantics to infer the compositions. Li \etal~\cite{li2023distilled} disentangle attributes and objects with reversed attention.
The second strategy directly predicts compositions by aligning images and textual labels in a shared space and searching for most similar compositions~\cite{xu2021relation,nagarajan2018attributes,saini2022disentangling,li2022siamese}. For example, Nagarajan \etal~\cite{nagarajan2018attributes} build a composition space by simulating all the visual changes of attributes performed on objects. Anwaar \etal~\cite{anwaar2022leveraging} improve composition learning by building a composition graph.  Recent approaches~\cite{lu2023decomposed,csp2023,wang2023hierarchical}, rooted in Vision-Language Models (VLM), also adopt either of the two strategies, utilizing pre-trained VLM encoders to better encode and align images and texts. For example, Nayak \etal adapt prompt in CLIP~\cite{radford2021learning} to fit the CZSL task. Lu \etal~\cite{lu2023decomposed} further boost the performance by soft prompts and disentangling strategy. In this study, we unify two strategies and a fixed VLM backbone within a single model and further enhance the model with specificity-refined attribute learning and specificity-based composition filtering.

\paragraph{Vision-Language Model (VLM).}

VLM~\cite{rebuffi2017learning,rebuffi2018efficient,mahabadi2021parameter,radford2018improving,li2021prefix} demonstrates remarkable potential in addressing various vision and language tasks, such as visual question answering~\cite{parelli2023clip} and image captioning~\cite{zhang2022distinctive}. Recent approaches have enhanced VLM's compatibility with downstream tasks by incorporating small adapters into the network or customizing prompt engineering. Small adapters~\cite{rebuffi2017learning,rebuffi2018efficient,mahabadi2021parameter,lu2023decomposed} refer to additional layers added to VLMs. They can boost VLMs' performance on downstream tasks with minimal network parameters fine-tuned. For instance, Mahabadi \etal~\cite{mahabadi2021parameter} introduce additional layers in each transformer block for multi-task fine-tuning. Meanwhile, prompt engineering~\cite{wang2023hierarchical,csp2023,lu2023decomposed,gu2021ppt,li2021prefix} enhance large pre-trained models like CLIP~\cite{radford2021learning} and GPT~\cite{brown2020language} by changing prompt guidance. Prompts can be static text or learnable word embeddings, aiming to help models quickly adapt to new tasks with little or none retraining. For instance, Wang \etal~\cite{wang2023hierarchical} utilize hierarchical prompts to enhance CLIP's performance in CZSL. In our model, we investigate both adapters and prompt engineering to improve the performance.

\begin{figure*}
\centering
  \includegraphics[width=\linewidth]{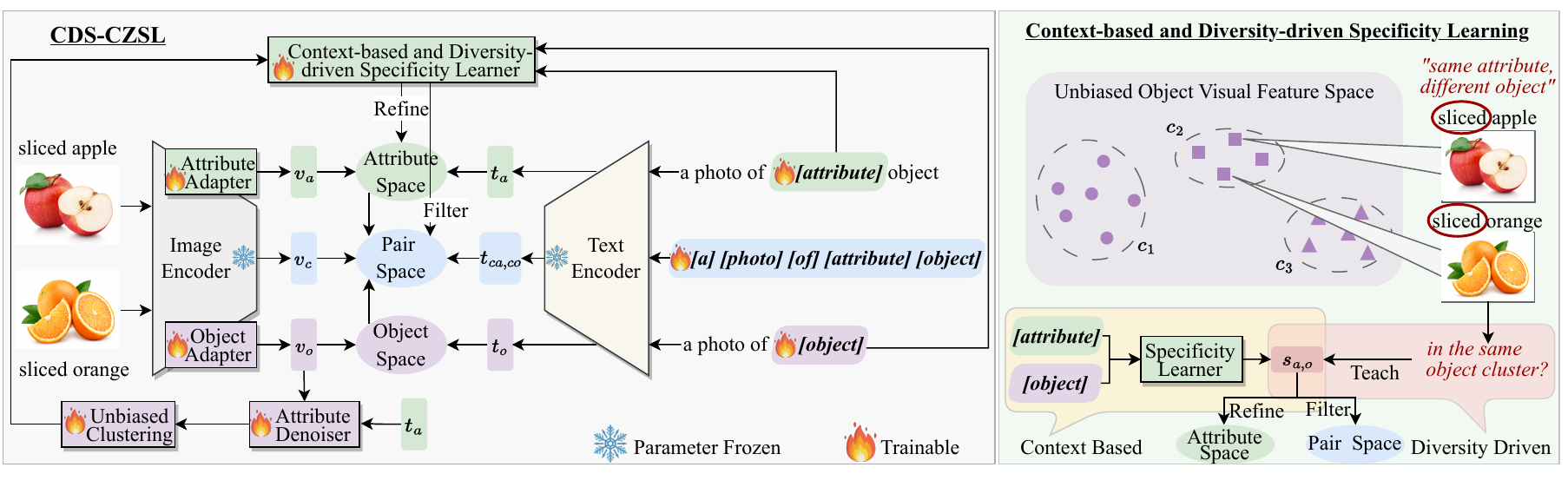}
    \caption{CDS-CZSL Overview and process of the context-based and diversity-driven specificity learning.}
    \label{fig:model}
\end{figure*}

\section{Method}
\label{sec:method}

\textbf{Problem definitions and notations.} In CZSL, images are modeled as compositions of primitives, i.e., attributes $ a \in \mathcal{A} $ and objects $ o \in \mathcal{O} $. This leads to a label space $ Y = \mathcal{A} \times \mathcal{O} $, capturing all possible attribute-object combinations. Within this space, we distinguish between seen compositions $Y^{\mathcal{S}}$ and unseen compositions $Y^{\mathcal{U}}$, with $ Y^{\mathcal{S}} \cap Y^{\mathcal{U}} = \emptyset $.   
During training, we have access to data from seen compositions: $ \mathcal{S} = \{ (x, y) | x \in X^{\mathcal{S}}, y \in Y^{\mathcal{S}} \} $, where each image $ x $ is labeled with an attribute-object pair $ y = (a, o) $. Then, during testing, the model needs to predict labels for images from both seen and unseen compositions. Depending on the scope of the output label space, we have two test settings: CW-CZSL and OW-CZSL \cite{mancini2021open}. In CW, $ Y^{\mathcal{U}} $ is given as prior knowledge, and the testing label space is restricted to $ y \in Y^{\mathcal{S}} \cup Y^{\mathcal{U} }$, while in OW, the output space expands to all potential attribute-object pairs, i.e., $ y \in Y $.

\textbf{Overview.} To tackle CZSL, we propose the CDS-CZSL framework. It consists of three distinct branches, tailored for predictions within attribute, object, and pair spaces.  Through pre-trained image \(f_{i}\) and text \(f_{t}\) encoders, we project images and composition labels into the common pair space for direct composition predictions. For attribute and object predictions, we enhance their visual representations with an attribute adapter \(f_{a}\) and an object adapter \(f_{o}\), both anchored on the image encoder \(f_{i}\), and improve their semantic understandings with two specialized prompts. We further design a context-based and diversity-driven specificity learner \(f_{s}\) to 1) refine the attribute learning process, thus enabling the prioritization of specific attributes, and 2) filter out undesired compositions in the OW pair space.


\subsection{Composition-wise Learning}

As mentioned before, contextuality is crucial for  CZSL~\cite{misra2017red,nagarajan2018attributes} due to the diverse appearances of attributes and objects across varied compositions. Addressing this, we present the composition-wise learning branch. The main idea behind this method is to create a unified space to which both images and composition labels are projected. Within this space, a similarity search is conducted to find the most compatible visual and semantic representations. In line with prior research~\cite{lu2023decomposed,csp2023}, we utilize transformer-based encoders for projecting both visuals and labels. Notably, these encoders, pre-trained using Contrastive Language Image Pre-Training (CLIP)~\cite{radford2021learning}, are kept frozen throughout our training.


To effectively harness the power of CLIP for our CZSL task, we reformat composition labels into structured natural language prompts like \underline{\textit{[a] [photo] [of] [\textbf{attribute}] [\textbf{object}]}}.
The prompt is fully soft in that each word in the prompt is modeled as learnable parameters $\theta_{c}=\{w_{0},w_{1},w_{2},w_{ca},w_{co}\}$, where the first three parameters represent prefix word embeddings, while the last two represent attribute and object word embeddings.
Using the text encoder $f_{t}$, these embeddings yield a text representation $t_{ca, co}$; in parallel, the image encoder $f_{i}$ processes the given image $x$ to produce the image representation $v_{c}$:
\begin{equation}
t_{ca, co}= f_{t}(\theta_{c}),v_{c} = f_{i}(x)
\end{equation}


Finally, we normalize both text and image representations using $\ell$2-normalization: $t_{ca, co}=\frac{t_{ca, co}}{\left \| t_{ca, co} \right \| }$ and $v_{c}=\frac{v_{c}}{\left \| v_{c} \right \| }$, and obtain the probability for class $y=(ca,co)$  by:

\begin{equation}
\resizebox{0.9\linewidth}{!}{$
    p(y=(ca,co)|x)=\frac{exp(v_{c}\cdot t_{ca, co}/ \tau  )}{\sum_{(\hat{ca},\hat{co})\in Y^{\mathcal{S}}}exp(v_{c}\cdot t_{\hat{ca}, \hat{co}}/ \tau  )}$}
\end{equation}
where $\tau$ is a temperature parameter from CLIP.
During training, we optimize $\theta_{c}$ to produce better composition-level representations by minimizing the cross-entropy loss:

\begin{equation}
    \min_{\theta_{c}}\mathcal{L}_{\mathit{base}}=-\frac{1}{\left | \mathcal{S} \right | } \sum_{(x,y)\in \mathcal{S}}\log p(y|x)
\end{equation}
where, $\left | \mathcal{S} \right | $ denotes the number of instances in the seen set.


Though composition-wise learning captures the compositional contextuality and excels in extracting composition-level representations, it struggles when generalized to OW seniors~\cite{li2023distilled,liu2021rethink}. Hence, we propose primitive-wise learning for distinct attribute and object learning.

\subsection{Primitive-wise Learning}

In this section, we introduce our primitive-wise learning branches. Distinct from composition-level insights of the composition-wise branch, primitive-wise learning ensures a deeper understanding of individual attributes and objects. It also lays the foundation for subsequent specificity learning.


For semantic primitive understanding, we adopt the following prompts: \underline{\textit{a photo of [\textbf{attribute}] object}} for attribute learning and \underline{\textit{a photo of [\textbf{object}]}} for object learning. An additional suffix (\textit{object}) is inserted in the attribute prompt to ensure sentence completeness. We denote the prompts as $\theta_{a}=\{e_{0},e_{1},e_{2},w_{a},e_{3}\}$ for attributes and $\theta_{o}=\{e_{0},e_{1},e_{2},w_{o}\}$ for objects. Unlike the full soft composition prompt $\theta_{c}$, in $\theta_{a}$ and $\theta_{o}$, only the attribute and object word embedding $w_{a}$ and $w_{o}$ are learnable, and the remainings are fixed during training. This design ensures these two branches focus exclusively on attribute/object information. Primitive-level text representations are then obtained by:

\begin{equation}
    t_{a}=f_{t}(\theta_{a}), t_{o}=f_{t}(\theta_{o})
\end{equation}

For image representations, a Multi-Head Self-Attention layer~\cite{vaswani2017attention} is introduced after the image encoder, functioning as the attribute/object adapter. Distinct adapters, attribute adapter $f_{a}$ and object adapter $f_{o}$, are employed, allowing the derivation of unique visual representations for attributes and objects respectively:

\begin{equation}
    v_{a}=f_{a}(f_{i}(x)), v_{o}=f_{o}(f_{i}(x))
\end{equation}

Finally, we normalize the representations $t_{a}=\frac{t_{a}}{\| t_{a} \|}$, $t_{o}=\frac{t_{o}}{\| t_{o} \|}$, $v_{a}=\frac{v_{a}}{\| v_{a} \|}$, and $v_{o}=\frac{v_{o}}{\| v_{o} \|}$, and compute attribute and object probabilities as follows:

\begin{equation}
    p(a|x)=\frac{exp(v_{a}\cdot t_{a}/ \tau)}{\sum_{\hat{a}\in \mathcal{A}}exp(v_{a}\cdot t_{\hat{a}}/ \tau)}
    \label{eq:att}
\end{equation}

\begin{equation}
    p(o|x)=\frac{exp(v_{o}\cdot t_{o}/ \tau)}{\sum_{\hat{o}\in \mathcal{O}}exp(v_{o}\cdot t_{\hat{o}}/ \tau)}
\end{equation}

The learned visual representations for objects $v_{o}$ and predicted attribute probabilities $p(a|x)$ are then fed into our Context-based and Diversity-driven Specificity Learner to learn specificity in attributes.

\subsection{Context-based and Diversity-driven Specificity Learning}

To determine attribute specificity, our approach is based on the intuition that an attribute's specificity inversely relates to the range of object clusters it describes. Note that we consider an attribute's descriptive diversity across object clusters rather than individual categories. This is because, taking \cref{fig:intro-b} as an example,  both `Sliced' and `Red' apply to several object, but `Sliced' is more specific as it describes similar fruits, such as apples and strawberries, while `Red' is less specific because it applies to a broader range of unrelated objects, like apples and cars. 

When clustering objects, attributes entangled in their representations can lead to biased clusters—`Red Car' and `Red Apple' might cluster together due to the shared attribute. To address this, we apply an attribute denoising step before clustering, ensuring that objects are grouped by their inherent characteristics rather than by shared attributes.


\textbf{Attribute Denoiser.} We design the Attribute Denoiser following the idea in ~\cite{liu2023pami} that image representation $v_{o}$ can be seen as attribute-denoised if we cannot infer the attribute from $v_{o}$. Thus, we first infer the attribute using  $v_{o}$ by:

\begin{equation}
    p(a|v_{o})=\frac{exp(v_{o}\cdot t_{a}/ \tau  )}{\sum_{\hat{a}\in \mathcal{A}}exp(v_{o}\cdot t_{\hat{a}}/ \tau  )}
\end{equation}
Then we introduce a denoising loss $\mathcal{L}_{\mathit{den}}$ calculated by the mean square error to guide the predicted attribute probability towards a uniform distribution:
\begin{equation}
    \min_{f_{o}}\mathcal{L}_{\mathit{den}}=\frac{1}{|\mathcal{A}|}\sum_{a\in \mathcal{A}}(\frac{1}{|\mathcal{A}|}-p(a|v_{o}))^{2}
\end{equation}
Note that we detach the gradient of $t_{a}$ during the calculation of $\mathcal{L}_{\mathit{den}}$ to prevent the denoising loss from compromising the precision of the text representation of the attribute. 



\textbf{Diversity-driven Specificity Learner.}
After achieving unbiased object representations $v{_o}$, we aim to cluster $v{_o}$ to infer attributes' descriptive diversity. Directly clustering the entire dataset every time $v_{o}$ is updated during training would require storing all instances of $v_{o}$ and recalculating clusters after each update, a process both computationally and temporally prohibitive. To address this, we opt for batch-level cluster updates (using K-Means clustering~\cite{scikit-learn}), and infer diversity by each time randomly selecting pairs of images $x$ and $\hat{x}$  that share attributes but differ in objects. And then, if their object representations $v_{o}$ and $\hat{v}_{o}$ are in the same cluster, it indicates that objects linked to this attribute share visual similarities, resulting in low diversity. Conversely, different clusters signify high diversity. The probability of co-clustering serves as an indirect measure of attributes' descriptive diversity across the dataset. 




We then estimate the specificity as binary: it is 1 if $v_{o}$ and $\hat{v}_{o}$ cluster together, indicating low descriptive diversity of attribute $a$ and thus high specificity, and 0 if they do not co-cluster, indicating low specificity. For an image $x$ with the label pair $(a,o)$, the specificity $\mathit{s}_{a,o}$ is quantified as follows:
\begin{equation}
    \mathit{s}_{a,o}=\begin{cases}
  1 & \text{ if } \Gamma (v_{o})=\Gamma (\hat{v}_{o}) \\
  0 & \text{ if } \Gamma (v_{o})\neq \Gamma (\hat{v}_{o})
\end{cases}
\end{equation}
where $\Gamma$ denotes the K-Means algorithm. 

However, specificity is also context-dependent, as discussed in \cref{sec:intro}. Therefore, we do not use $\mathit{s}_{a,o}$ directly as the specificity. Instead, we introduce a specificity learner, $f_{s}$, which employs word embeddings $w_{a}$ and $w_{o}$ to predict specificity based on context and use $\mathit{s}_{a,o}$ as the target of $f_{s}$. This approach ensures that our specificity learner, $f_{s}$, is context-based and guided by diversity. We employ the cross-entropy loss function to optimize $f_{s}$:

\begin{equation}
    \min_{f_{s}}\mathcal{L}_{\mathit{div}}=-\frac{1}{\left | \mathcal{S} \right | } \sum_{(x,y)\in \mathcal{S}}\log p(\mathit{s}_{a,o}|x)
\end{equation}

\textbf{Specificity-refined Primitive Learning.} To encourage the model to prioritize attributes with higher specificity in its predictions, we introduce specificity predicted by $f_{s}$ as a penalty term. Then, the specificity-refined attribute prediction and the specificity-refined  primitive loss $\mathcal{L}_{prim}$ can be expressed as follows:
\begin{equation}\label{a_o_prediction}
    p'(a|x)= p(a|x)-f_{s}(w_{a},w_{o})/\gamma 
\end{equation}
\begin{equation}
\min_{w_{a},w_{o},f_{a},f_{o}}\mathcal{L}_{\mathit{prim}}=-\frac{1}{\left | \mathcal{S} \right | } \sum_{(x,y)\in \mathcal{S}}[\log p'(a|x)+\log p(o|x)]
\end{equation}
where, $\gamma=\frac{1}{|\mathcal{A}|}$ adjusts the range of the penalty term.

During training, a larger specificity penalty term results in a lower probability of the refined attribute prediction $p'(a|x)$. Then, to minimize $\mathcal{L}_{prim}$, the model leans towards assigning a higher probability to  $p(a|x)$ (\cref{eq:att}) to remedy the penalty term. By optimizing   $\mathcal{L}_{prim}$, this specificity penalty term amplifies $p(a|x)$ if $(a,o)$ exhibits high specificity. Consequently, when we remove this penalty term during testing, the probability $p(a|x)$ is elevated for attributes with high specificity, increasing the likelihood of predicting specific attribute $a$. Conversely, if the attribute $a$ is general, its smaller specificity penalty terms during training lead to lower prediction probabilities for $a$ when we remove the penalty terms during testing.

\subsection{Refined Prediction}


During training, we utilize the specificity-refined attribute and object probabilities to enhance the base model, yielding the fused prediction of our model:
\begin{equation}
\begin{split}
    p'(y|x)&=\alpha p(y=(ca,co)|x)+(1-\alpha) p'(y=(a,o)|x)\\
    &=\alpha p(y=(ca,co)|x)+(1-\alpha) p'(a|x)p(o|x)
\end{split}
\end{equation}

The refined prediction loss for our model is given by:
\begin{equation}
    \min_{\theta_{c},w_{a},w_{o},f_{a},f_{o}}\mathcal{L}_{\mathit{refine}}=-\frac{1}{\left | \mathcal{S} \right | } \sum_{(x,y)\in \mathcal{S}}\log p'(y|x)
\end{equation}

\subsection{Training and Inference}
\textbf{Training Objectives.} To ensure that specificity learning does not affect the optimization of representations, our training process is divided into two phases: representation learning and specificity learning. The representation learner encompasses adapters and learnable word embeddings, denoted as $\theta_{c}$, $w_{a}$, $w_{o}$, $f_{a}$, and $f_{o}$. The specificity learner is represented as $f_{s}$. The overall training loss is defined as follows:
\begin{equation}
\min_{\theta_{c},w_{a},w_{o},f_{a},f_{o}}\mathcal{L}_{\mathit{base}}+\mathcal{L}_{prim}+\mathcal{L}_{den}+\mathcal{L}_{\mathit{refine}};\min_{f_{s}}\mathcal{L}_{\mathit{div}}
\end{equation}
Initially, we optimize the representation learner exclusively for one epoch, ensuring reliable representations for subsequent specificity learning. All object image representations are retained as initial inputs for the K-Means clustering. Subsequently, we iteratively optimize the representation learner, K-means kernel, and specificity learner in a batch-wise manner.

\textbf{Inference.} During testing,  we exclude the penalty term in \cref{a_o_prediction} to obtain attribute probability (\cref{eq:att}) and fuse primitive predictions and composition predictions as the final results:
\begin{equation}
y'=\argmax_{y\in Y^{\mathcal{T}}}\alpha p(y=(ca,co)|x)+(1-\alpha) p(y=(a,o)|x)
\end{equation}
where, $Y^{\mathcal{T}}=Y^{\mathcal{S}}\cup Y^{\mathcal{U}}$ in the CW-CZSL, and $Y^{\mathcal{T}}=\mathcal{A}\times \mathcal{O}$ in the OW-CZSL.

Additionally, we employ specificity to filter out less likely pairs, reducing the search space for compositions in the open world. This strategy is not used in the closed-world scenario where all pairs are feasible. Specifically, we introduce two thresholds to retain compositions with moderate specificity, such that $Y^{\mathcal{T}}=\{y=(a,o):s_{a,o}\in [t_{l},t_{h}]\}$, excluding compositions that are overly specific or overly general, where $t_{l}< t_{h}$.

\begin{table}[]
\small
\centering
\setlength{\tabcolsep}{3pt} 
\begin{tabular}{lcc|cc|ccc}
\toprule
           &     &        & \multicolumn{2}{c|}{Training} & \multicolumn{3}{c}{Testing}                \\
Dataset    & a   & o  &sp     & i            & sp & up & i    \\
\midrule
MIT-States & 115 & 245   & 1262          & 30k          & 400  & 400(27k)    & 13k                  \\
UT-Zappos  & 16  & 12      & 83            & 23k          & 18   & 18(109)     & 3k                 \\
C-GQA      & 413 & 674  & 5592          & 27k          & 888 & 923(272k)   & 5k     \\
\bottomrule
\end{tabular}
\caption{Statistics of datasets: a, o, i, sp, and up are the number of attributes, objects, images, seen pairs, and unseen pairs. Numbers in brackets are the unseen search space in OW.  }
\label{tab:datasets}
\end{table}

\begin{table*}
\small
\centering
\begin{tabular}{c|l|cccc|cccc|cccc}
\toprule
&\multirow{2}*{Closed-World} & \multicolumn{4}{c|}{MIT-States}                  & \multicolumn{4}{c|}{UT-Zappos}                 & \multicolumn{4}{c}{C-GQA}\\
& & ~S~& ~U~& HM& AUC & ~S~ & ~U~& HM & AUC& ~S~& ~U~& HM&AUC\\ \midrule
\multirow{7}*{\rotatebox{90}{\textbf{w/o CLIP}}} &CGE~\cite{naeem2021learning} & 32.8 & 28.0 & 21.4 & 6.5 & \textcolor{blue}{64.5} & 71.5 & \textbf{60.5} & 33.5 & 31.4 & 14.0 & 14.5 & 3.6 \\
&CompCos~\cite{mancini2021open} & 25.3 & 24.6 & 16.4 & 4.5 & 59.8 & 62.5 & 43.1 & 28.7 & 28.1 & 11.2 & 12.4 & 2.6 \\
&Co-CGE~\cite{mancini2022learning} & 32.1 & 28.3 & 20.0 & 6.6 & 62.3 & 66.3 & 48.1 & 33.9 & 33.3 & 14.9 & 14.4 & 4.1 \\
&SCEN~\cite{li2022siamese} & 29.9 & 25.2 & 18.4 & 5.3 & 63.5 & 63.1 & 47.8 & 32.0 & 28.9 & 25.4 & 17.5 & 5.5 \\
W/O&CANet~\cite{wang2023learning} & 29.0 & 26.2 & 17.9 & 5.4 & 61.0 & 66.3 & 47.3 & 33.1 & 30.0 & 13.2 & 14.5 & 3.3 \\
&CoT~\cite{kim2023hierarchical} & 34.8 & 31.5 & 23.2 & 7.8 & - & - & - & - & 34.0 & 18.8 & 17.5 & 5.1 \\
&ADE~\cite{hao2023learning} & - & - & - & - & 63.0 & 64.3 & 51.1 & 35.1 & 35.0 & 17.7 & 18.0 & 5.2 \\
\midrule
\multirow{5}*{\rotatebox{90}{\textbf{w CLIP}}}& CLIP~\cite{radford2021learning} & 30.2 & 46.0 & 26.1 & 11.0 & 15.8 & 49.1 & 15.6 & 5.0  & 7.5  & 25.0 & 8.6  & 1.4  \\
&CLIP-based Co-CGE~\cite{mancini2022learning} & 46.7 & 45.9 & 33.1 & 17.0 & 63.4 & 71.3 & 49.7 & \textcolor{blue}{36.3} & 34.1 & 21.2 & 18.9 & 5.7  \\
&CoOP~\cite{zhou2022learning} & 34.4 & 47.6 & 29.8 & 13.5 & 52.1 & 49.3 & 34.6 & 18.8 & 20.5 & 26.8 & 17.1 & 4.4 \\
&CSP~\cite{csp2023} & 46.6 & 49.9 & 36.3 & 19.4 & 64.2 & 66.2 & 46.6 & 33.0 & 28.8 & 26.8 & 20.5 & 6.2 \\
&HPL~\cite{wang2023hierarchical} & \textcolor{blue}{47.5} & 50.6 & \textcolor{blue}{37.3} & 20.2 & 63.0 & 68.8 & 48.2 & 35.0 & 30.8 & 28.4 & 22.4 & 7.2 \\
&DFSP~\cite{lu2023decomposed} & 46.9 & \textcolor{blue}{52.0} & \textcolor{blue}{37.3} & \textcolor{blue}{20.6} & \textbf{66.7} & \textcolor{blue}{71.7} & 47.2 & 36.0 & \textcolor{blue}{38.2} & \textcolor{blue}{32.0} & \textcolor{blue}{27.1} & \textcolor{blue}{10.5} \\
\cmidrule(lr){2-14}
&CDS-CZSL (ours) & \textbf{50.3} & \textbf{52.9} & \textbf{39.2} & \textbf{22.4} & 63.9 & \textbf{74.8} & \textcolor{blue}{52.7} & \textbf{39.5} & \textbf{38.3} & \textbf{34.2} & \textbf{28.1} & \textbf{11.1} \\
\bottomrule
\end{tabular}
\caption{Model performance in CW. We use `w' and `w/o' to distinguish models adopting CLIP as visual and language encoders or not. The best results are in \textbf{bold}. The second best results are in \textcolor{blue}{blue}.}
\label{tab:close}
\end{table*}

\section{Experiments}
\label{exp}

\subsection{Experiment Settings}

\textbf{Datasets and evaluation metrics.} 
We evaluate our model on three benchmark datasets: 1) MIT-States ~\cite{isola2015discovering} features natural objects with diverse attributes. Its relatively noisy data and fine-grained attributes make it challenging to learn; 
2) UT-Zappos is a specialized, small-scale dataset centered on footwear, encompassing a limited set of 16 attributes and 12 objects;
3) C-GQA stands out with its expansive vocabulary, 413 attributes and 674 objects. Such breadth presents a formidable challenge, particularly in OW scenarios. 
Datasets are split into seen and unseen compositions following the split in previous works~\cite{purushwalkam2019task,naeem2021learning} to ensure fair comparisons. The splits are based on generalized CZSL setting~\cite{naeem2021learning} to ensure both seen and unseen compositions appear during testing. Split details are in \cref{tab:datasets}.

We evaluate our model using the protocol from \cite{purushwalkam2019task,mancini2021open}. Varied biases are added to unseen pairs to adjust testing results, during which the best-seen accuracy (S), best-unseen accuracy (U), best harmonic mean (HM), and Area Under the Curve (AUC) of the HM-bias curve are recorded as performance indicators. Among these, AUC is the core metric as it evaluates the model comprehensively.

\textbf{Implement details.} We follow prior practices~\cite{csp2023,lu2023decomposed} to adopt CLIP~\cite{radford2021learning} as our image/text encoder. Object and attribute adapters are one-layer Multi-head Attention. The specificity learner is a four-layer Fully-Connected Network( FCN). We use K-means to cluster object representations, and the clustering is batch-wise to save computation costs. 
The cluster number is calculated using the base-2 logarithm of the object classes, resulting in numbers of 8, 4, and 10 for three datasets, respectively. 
The model is trained end-to-end with Adam optimizer~\cite{kingma2014adam}. Hyper-parameters, such as learning rate and fusion weight ($\alpha$), are determined based on validation set performance.   The supplementary provides codes with detailed parameters.

\begin{table*}
\small
\centering
\begin{tabular}{c|l|cccc|cccc|cccc}
\toprule
&\multirow{2}*{Open-World} & \multicolumn{4}{c|}{MIT-States}                  & \multicolumn{4}{c|}{UT-Zappos}                 & \multicolumn{4}{c}{C-GQA}\\
& & ~S~& ~U~& HM& AUC & ~S~ & ~U~& HM & AUC& ~S~& ~U~& HM&AUC\\ \midrule
\multirow{7}*{\rotatebox{90}{\textbf{w/o CLIP}}}  &CGE~\cite{naeem2021learning}  & 32.4 & 5.1 & 6.0 & 1.0 & 61.7 & 47.7 & 39.0 & 23.1 & 32.7 & 1.8 & 2.9 & 0.47 \\
&CompCos~\cite{mancini2021open}  & 25.4 & 10.0 & 8.9 & 1.6 & 59.3 & 46.8 & 36.9 & 21.3 & 28.4 & 1.8 & 2.8 & 0.39 \\
&Co-CGE~\cite{mancini2022learning} & 30.3 & 11.2 & 10.7 & 2.3 & 61.2 & 45.8 & 40.8 & 23.3 & 32.1 & 3.0 & 4.8 & 0.78 \\
&KG-SP~\cite{karthik2022kg} & 28.4 & 7.5 & 7.4 & 1.3 & 61.8 & 52.1 & 42.3 & 26.5 & 31.5 & 2.9 & 4.7 & 0.78 \\
&SAD-SP~\cite{liu2023pami} & 29.1 & 7.6 & 7.8 & 1.4 & 63.1 & 54.7 & 44.0 & 28.4 & 31.0 & 3.9 & 5.9 & 1.00 \\
&DRANet~\cite{li2023distilled} & 29.8 & 7.8 & 7.9 & 1.5 & \textcolor{blue}{65.1} & 54.3 & 44.0 & 28.8 & 31.3 & 3.9 & 6.0 & 1.05 \\
&ADE~\cite{hao2023learning} & - & - & - & - & 62.4 & 50.7 & 44.8 & 27.1 & 35.1 & 4.8 & 7.6 & 1.42 \\
\midrule
\multirow{5}*{\rotatebox{90}{\textbf{w CLIP}}}  &
CLIP~\cite{radford2021learning}& 30.1 & 14.3 & 12.8 & 3.0  & 15.7 & 20.6 & 11.2 & 2.2  & 7.5  & 4.6  & 4.0  & 0.27  \\
&CLIP-based Co-CGE~\cite{mancini2022learning}& 38.1 & \textcolor{blue}{20.0}   & 17.7 & 5.6  & 59.9 & 56.2 & \textcolor{blue}{45.3} & 28.4 & 33.2 & 3.9  & 5.3  & 0.91\\
&CoOP~\cite{zhou2022learning} & 34.6 & 9.3 & 12.3 & 2.8 & 52.1 & 31.5 & 28.9 & 13.2 & 21.0 & 4.6 & 5.5 & 0.70 \\
&CSP~\cite{csp2023} & 46.3 & 15.7 & 17.4 & 5.7 & 64.1 & 44.1 & 38.9 & 22.7 & 28.7 & 5.2 & 6.9 & 1.20 \\
&HPL~\cite{wang2023hierarchical} & 46.4 & 18.9 & \textcolor{blue}{19.8} & \textcolor{blue}{6.9} & 63.4 & 48.1 & 40.2 & 24.6 & 30.1 & 5.8 & 7.5 & 1.37 \\
&DFSP~\cite{lu2023decomposed} & \textcolor{blue}{47.5} & 18.5 & 19.3 & 6.8 & \textbf{66.8} & \textcolor{blue}{60.0} & 44.0 & \textcolor{blue}{30.3} & \textbf{38.3} &  \textcolor{blue}{7.2} & \textcolor{blue}{10.4} & \textcolor{blue}{2.40} \\
\cmidrule(lr){2-14}
&CDS-CZSL w filtering (ours) & \textbf{49.4} & \textbf{21.8} & \textbf{22.1} & \textbf{8.5} &\textbf{64.7}	&\textbf{61.3}	&\textbf{48.2}	&\textbf{32.3}
 & \textcolor{blue}{37.6} & \textbf{8.2} & \textbf{11.6} & \textbf{2.68} \\
\bottomrule
\end{tabular}
\caption{Model performance in OW. `w filtering' indicates that the reported CDS-CZSL uses specificity to filter compositions.}
\label{tab:open}
\end{table*}

\subsection{Comparisons with SOTAs}
We compare our CDS-CZSL with most recent CZSL methods~\cite{mancini2021open,mancini2022learning,li2022siamese,wang2023learning,kim2023hierarchical,karthik2022kg,liu2023pami,li2023distilled,hao2023learning,radford2021learning,mancini2022learning,zhou2022learning,csp2023,lu2023decomposed} in both CW and OW settings. Given the same data splits and evaluation metrics, the reported performances from the original publications are directly used for competitors. The results of all CLIP-based methods are run with ViT-L/14 with CLIP fixed during training.

The CW results, presented in \cref{tab:close}, reveal that CDS-CZSL achieves the best results on all datasets. Specifically, it yields improvements of 1.8\%, 3.2\%, and 0.6\% in AUC over the second-best methods on three datasets. It also attains considerable gains in HM on MIT-States and C-GQA, with increases of 1.9\% and 1\%, respectively. Although CGE outperforms it in HM on UT-Zappos, the overall performance of CDS-CZSL surpasses that of CGE, particularly in AUC, with a 6\% improvement.

Comparing methods with and without CLIP as backbones, we observe that CLIP-based methods possess higher performance ceilings. However, our enhancements are not solely due to CLIP. Compared with other CLIP-based methods, our performance still prevails. Firstly, unlike CSP~\cite{csp2023} and DFSP~\cite{lu2023decomposed}, which only predict in the composition space, our approach also leverages separate learning of attributes and objects, thereby exhibiting better generalization capabilities. This is also proved by our model achieving the highest unseen accuracy (U) across all datasets. Secondly, we introduce context-based and diversity-driven specificity learning that prioritizes informative specific attributes, thus facilitating more accurate predictions.

The OW results, depicted in \cref{tab:open}, indicate a significant performance drop for all methods transitioning from CW to OW. Yet, CDS-CZSL continues to outperform in all criteria (bar the seen accuracy in C-GQA) across datasets. Particularly, it achieves improvements of 1.6\%, 2\%, and 0.28\% in AUC. While the absolute increments are modest compared to the CW scenario, the relative improvements are higher on MIT-States (23.2\% vs. 8.7\%) and C-GQA (11.7\% vs. 5.7\%), confirming the effectiveness of our model. In addition to the aforementioned reasons in CW analysis, our specificity-based filtering strategy also contributes to the improvements in the OW setting. The lower relative improvements on UT-Zappos for OW (6.7\%) compared to CW (8.8\%) may be due to its small-scale and fine-grained nature, making our specificity learning and filtering less effective. However, the model still benefits from our design of composition-wise and primitive-wise learning, hence outperforming other methods. Notably, our model achieves the best U in OW as well; this demonstrates that our model's specificity learning does not compromise the model's generalization ability to unseen compositions.

\begin{table}
\centering
\small 
\setlength{\tabcolsep}{3pt} 
\begin{tabular}{cl|cccc|cccc}
\toprule
 & & \multicolumn{4}{c|}{MIT-States} & \multicolumn{4}{c}{UT-Zappos} \\
 & & S & U & HM & AUC & S & U & HM & AUC \\
\midrule
\multirow{5}*{\rotatebox{90}{\textbf{CW}}} &
 SPM & 44.1 & 51.7 & 35.8 & 19.1 & 64.3	&66.3	&47.3	&33.8 \\
 &3branch & 47.1 & \textbf{53.2} & 37.5 & 21.1 & \textcolor{blue}{64.5} & 71.3 & 49.0 & 36.2 \\
 & CBS & \textcolor{blue}{50.0} & 52.5 & \textcolor{blue}{38.4} & \textcolor{blue}{22.0} & \textbf{65.1} & \textbf{75.0} & 52.5 & \textcolor{blue}{39.3} \\
  & DDS & 49.6 & 52.5 & 38.2 & 21.8 & 63.2 & 73.5 & \textcolor{blue}{52.2} & 38.6 \\
  &CDS & \textbf{50.3} & \textcolor{blue}{52.9} &  \textbf{39.2} &  \textbf{22.4} & 63.9 & \textcolor{blue}{74.8} & \textbf{52.7} & \textbf{39.5} \\
\midrule
\multirow{5}*{\rotatebox{90}{\textbf{OW}}} &
 SPM & 45.8 & 16.7 & 18.3 & 6.1 & 63.2 & 52.0 & 44.1 & 27.1 \\
 & 3branch & 45.0 & 21.1 & 20.6 & 7.4 & 61.9 & \textcolor{blue}{60.0} & 45.2 & 29.9 \\
 & CBS & 48.3 & \textbf{22.1} & \textcolor{blue}{21.8} & \textcolor{blue}{8.3} & 63.5 & 59.0 & \textcolor{blue}{47.9} & \textcolor{blue}{31.0} \\
 & DDS & \textbf{50.1} & 21.2 & 21.7 & \textcolor{blue}{8.3} & \textbf{66.8} & 56.0 & 45.4 & 29.5 \\
 & CDS & \textcolor{blue}{49.4} & \textcolor{blue}{21.8} & \textbf{22.1} & \textbf{8.5} & \textcolor{blue}{64.7}	& \textbf{61.3}	&\textbf{48.2}	&\textbf{32.3} \\
\bottomrule
\end{tabular}
\caption{Module ablation study. CDS denotes our CDS-CZSL with `-CZSL' removed to save space. }
\label{tab:ablation}
\end{table}

\begin{table*}[]
\small
\centering
\begin{tabular}{l|ccccc|ccccc}
\toprule
 & \multicolumn{5}{c|}{UT-Zappos} & \multicolumn{5}{c}{C-GQA} \\
& feasible pairs & S & U & HM & AUC & feasible pairs & S & U & HM & AUC  \\
\midrule
w/o filtering  & 192 & 64.7 & 60.3 & 47.9 & 32.0 &278362& - & - & - & - \\
w similarity-based filtering~\cite{lu2023decomposed} & 189 & 64.7 & 60.4 & 48.0 & 32.1 & 67293 & 38.2 & 8.0 & 10.9 & 2.61 \\
w our filtering & 176 & 64.7 &61.3 & 48.2 & 32.3
 & 30332 & 37.6 & 8.2 & 11.6 & 2.68 \\
\bottomrule
\end{tabular}
\caption{Ablation on filtering strategy.}
\label{tab:filtering}
\end{table*}

\begin{figure*}
\centering
  \includegraphics[width=0.9\linewidth]{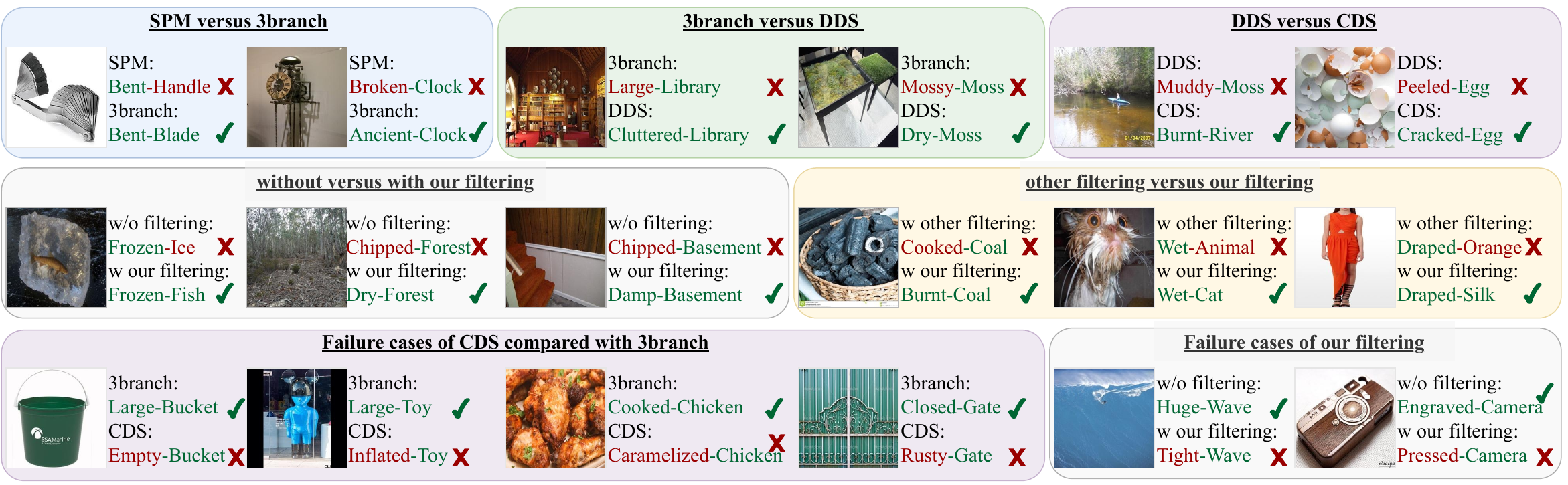}
    \caption{Qualitative Results of varying network structure (first row), changing filtering method (second row), and failure cases (third row). }
    \label{fig:exp_visual}
\end{figure*}

\subsection{Ablation Study}\label{sec ablation}
\textbf{Module ablation.} In our ablation study, we evaluate the efficacy of each component in our proposed CDS-CZSL by comparing it against four variants. The variant \textit{SPM} contains only the composition-wise learning branch, while \textit{3branch} extends \textit{SPM} by incorporating primitive-wise learning branches for both attributes and objects but without the specificity learning component. To assess the impact of specificity learning, we introduce \textit{CBS}, which integrates context-based specificity into \textit{3branch}, and \textit{DDS}, which incorporates diversity-driven specificity.

The results, presented in \cref{tab:ablation}, illustrate several key findings. Firstly, the comparison between SPM and CSP~\cite{csp2023}---which differs from SPM in terms of whether the prompt is fully learnable---highlights the efficiency gains afforded by fully learnable prompts within the composition-wise branch. Furthermore, {3branch} demonstrates an improvement over SPM, validating the importance of including primitive-wise learning. Significantly, our model outperforms HPL~\cite{wang2023hierarchical}, which also employs a three-branch structure. This proves the advantages of our unique prompt designs and the integration of adapters in primitive learning.

Introducing specificity learning through both CBS and DDS leads to further performance gains over 3branch, confirming the effectiveness of specificity.  The superior performance of CBS over DDS suggests that context-based adaptability provides additional advantages over the cluster-based, less-learnable diversity strategy of DDS. However, DDS's success indicates the importance of descriptive diversity in ascertaining attribute specificity.
Combining the strengths of both context and diversity, our complete model, CDS-CZSL, achieves the highest results in terms of HM and AUC. This validates that both context and descriptive diversity are vital components of specificity learning in CZSL.

\textbf{Filtering ablation.} To assess the effectiveness of our specificity-based filtering strategy in OW, we compare it against no filtering and similarity-based filtering~\cite{lu2023decomposed}. Filtering thresholds are chosen on validation sets. Due to GPU limitations, we excluded unfiltered results for C-GQA. The results, detailed in \cref{tab:filtering}, show that while both filtering reduces the search space and increases performance, our specificity-based approach retains fewer but more relevant pairs, leading to superior HM and AUC. This proves its ability to effectively discard both overly general and highly specific pairs, unlike the similarity-based method, which may preserve generic pairs and omit unique, specific ones. 

\subsection{Qualitative Results}

\textbf{Module ablation}. We study the qualitative results to explore the effects of varying network structures in the first row of \cref{fig:exp_visual}. Initially, SPM is misled by seen pairs (e.g., Bent-Handle), while 3branch overcomes this through primitive-wise learning but struggles with too-general pairs like Mossy-Moss. DDS learning corrects this, refining such pairs to more specific Dry-Moss. CDS-CZSL further improves DDS with contexts, as seen when Peeled-Egg is adjusted to Cracked-Egg. 

\textbf{Filtering}.  The efficacy of our filtering strategy is shown in the second row; it can exclude overly general pairs, such as Frozen-Ice, and too specific ones, like Chipped-Basement. This contrasts with similarity-based filtering that might settle for broader categories, resulting in predictions like Wet-Animal instead of the more precise Wet-Cat.


\textbf{Limitations and potentials}. The third row in \cref{fig:exp_visual} shows cases where CDS-CZSL fails compared to 3branch and non-filtered version. It occasionally predicts specific pairs unsupported by the image content—for instance, we cannot tell if the Large-Bucket is Empty. Moreover, our filtering method is not unmistakable, retaining specific yet invalid pairs such as Pressed-Camera. However, CDS-CZSL, while not always accurate, often provides more descriptive labels than the original—refining Cooked-Chicken to Caramelized-Chicken, for example. This suggests that our model may have the potential to aid in refining labels.

\section{Conclusion}
\label{sec:conclusion}
In this work, we propose a Context-based and Diversity-driven Specificity learning framework for Compositional Zero-Shot Learning (CDS-CZSL). We incorporate composition-wise and primitive-wise learning to capture compositional contextuality and enhance attribute/object learning simultaneously. We then design the context-based and diversity-driven specificity learning to prioritize specific attributes that are more informative and use the learned specificity to filter compositions in Open-World scenarios. Through comprehensive experiments, we demonstrate the effectiveness of our model and achieve SOTA performance on three datasets. 
Furthermore, we discuss the limitations and potentials of our model leaning towards specific pairs.
{
    \small
    \bibliographystyle{ieeenat_fullname}
    \bibliography{main}
}


\end{document}